\documentclass[isre,nonblindrev]{informs4}
\usepackage{hyperref}
\hypersetup{
    colorlinks = true,
    urlcolor = {blue},
    linkcolor = blue,
    citecolor = [rgb]{0,0.5,0.5}
}
\usepackage{cleveref}
\usepackage{array}
\usepackage{graphicx}
\usepackage[flushleft]{threeparttable}
\usepackage{footnote}
\usepackage{multirow}
\usepackage{booktabs}
\usepackage{soul,xcolor}
\usepackage{titlesec}
\usepackage{appendix}
\usepackage{verbatim}
\usepackage{arydshln}
\usepackage{ebgaramond}
\usepackage{pdflscape}
\usepackage{colortbl,hhline}
\usepackage{longtable}
\usepackage{arydshln}
\usepackage{subcaption}
\usepackage[sort,comma]{natbib}
\usepackage{amsmath, bm}
\usepackage{enumitem}

\DeclareMathAlphabet{\mathsfit}{T1}{\sfdefault}{\mddefault}{\sldefault}
\SetMathAlphabet{\mathsfit}{bold}{T1}{\sfdefault}{\bfdefault}{\sldefault}

\usepackage{bbm}

\usepackage[sectionbib]{bibunits}
\defaultbibliographystyle{informs2014} 
\defaultbibliography{0.main}

\DoubleSpacedXI

\usepackage{natbib}
 \bibpunct[, ]{(}{)}{,}{a}{}{,}%
 \def\bibfont{\small}%
\TheoremsNumberedThrough     
\ECRepeatTheorems
\EquationsNumberedThrough    
\MANUSCRIPTNO{MS-0000-0000.00}

\newcommand{\TL}[1]{{\color{red}Tian: #1}}

\titleformat{\subsubsection}
  {\normalfont\normalsize\bfseries}{\thesubsubsection}{1em}{}
\titlespacing*{\subsubsection}{0pt}{3.25ex plus 1ex minus .2ex}{0ex plus .2ex}

\begin{document}

\RUNTITLE{}

\TITLE{Visioning Human–Agentic AI Teaming: Continuity, Tension, and Future Research\thanks{The authors are named in alphabetical order. All authors contributed equally.}}

\ARTICLEAUTHORS{%
\AUTHOR{Bowen Lou}
\AFF{University of Southern California, \EMAIL{bowenlou@marshall.usc.edu}}
\AUTHOR{Tian Lu}
\AFF{Arizona State University, \EMAIL{lutian@asu.edu}}
\AUTHOR{T. S. Raghu}
\AFF{Arizona State University, \EMAIL{Raghu.Santanam@asu.edu}}
\AUTHOR{Yingjie Zhang}
\AFF{Peking University, \EMAIL{yingjiezhang@gsm.pku.edu.cn}}

}

\OneAndAHalfSpacedXI

\ABSTRACT{%
Artificial intelligence is undergoing a structural transformation marked by 
the rise of agentic systems capable of open-ended action trajectories, 
generative representations and outputs, and evolving objectives. These properties introduce structural uncertainty into human–AI teaming (HAT), including uncertainty about behavior trajectories, epistemic grounding, and the stability of governing logics over time. Under such conditions, alignment cannot be secured through agreement on bounded outputs; it must be continuously sustained as plans unfold and priorities shift. We advance Team Situation Awareness (Team SA) theory, grounded in shared perception, comprehension, and projection, as an integrative anchor for this transition. While Team SA remains analytically foundational, its stabilizing logic presumes that shared awareness, once achieved, will support coordinated action through iterative updating. Agentic AI challenges this presumption. Our argument unfolds in two stages: first, we extend Team SA to reconceptualize both human and AI awareness under open-ended agency, including the sensemaking of projection congruence across heterogeneous systems. Second, we interrogate whether the dynamic processes traditionally assumed to stabilize teaming in relational interaction, cognitive learning, and coordination and control continue to function under adaptive autonomy. By distinguishing continuity from tension, we clarify where foundational insights hold and where structural uncertainty introduces strain, and articulate a forward-looking research agenda for HAT. We 
find that under open-ended agency, Team SA processes can produce the opposite of 
their intended effects: relational legitimacy may rest on epistemic fragility, 
iterative updating may amplify rather than correct divergence, and shared 
awareness may coexist with substantive loss of oversight. The central challenge of HAT is not whether humans and AI can agree in the moment, but whether they can remain aligned as futures are continuously generated, revised, enacted, and governed over time. 
}%


\maketitle
\setcounter{page}{0}
\thispagestyle{empty} 

\newpage
\DoubleSpacedXI

\section{Introduction}
\label{sec:intro}

Artificial intelligence (AI) is entering a structural transformation in how the underlying models represent the world, handle uncertainty, and participate in organized work. For decades, AI research and practice operated under a stabilizing assumption that the systems were task-specific, bounded in their uncertainty, and subordinate to direct human oversight. Symbolic AI and earlier machine learning (ML) systems largely fit this assumption profile. Their outputs were circumscribed, their failure modes were characterizable, and their decision policies were stationary unless a human deliberately intervened. A new class of systems has disrupted this premise: \textbf{agentic AI}. Enabled by large language models (LLMs) and increasingly autonomous architectures, agentic systems operate as goal-directed entities that \textit{reason}, \textit{plan}, \textit{act}, and \textit{reflect} over extended horizons, exhibiting a meaningful form of agency that allows their behavior to unfold over time rather than being fully specified at the moment of deployment \citep{mckinsey2023economic, mckinsey2024why, alavi2024genairedefine,dennis2023ai,durante2024agent,ng2023how}. This fundamentally less predictable operating mode suggests that existing theories, frameworks, and design principles developed for human-AI collaboration under bounded conditions may no longer be sufficient.  

What distinguishes agentic AI from traditional AI is not merely improved performance, but its open-ended agency, namely, the capacity to select, revise, and evolve actions, representations, and even objectives in ways that are neither fully specifiable ex ante nor fully verifiable ex post. 
This creates structural uncertainty along three dimensions of externally relevant behavior: 
what it does (action policy), what it produces (outputs and representations), and what governs it (the evolving configuration of objectives, constraints, and capabilities). 
Together, these map to uncertainty in trajectories, representations, and governing regimes that shape how action unfolds and remains accountable across episodes \citep{abouali2025agentic}. 
Accordingly, we distinguish three analytically separable forms of open-ended agency: 
First, \textbf{open-ended action trajectories (trajectory uncertainty)}. Agentic AI can autonomously choose, sequence, and revise multi-step actions, often involving tool use, intermediate subgoals, delegation, and replanning. This form of uncertainty operates within episodes and unfolds as action is executed and cannot be fully resolved before the trajectory completes. The operative path is therefore not fixed at invocation but emerges during execution as intermediate states and feedback redirect the plan. Uncertainty lies in the unfolding course of action: what the agent initiates, when it changes course, and how intervention is possible midstream \citep{huang2025path}.
Second, \textbf{open-ended representations and outputs (epistemic uncertainty)}. Agentic AI generates fluent explanations, ``facts,'' rationales, and artifacts (e.g., summaries, code, plans) whose epistemic status can be difficult to establish \citep{farquhar2024detecting,ling2024uncertainty}. This form of uncertainty attaches to any given output at a moment and is present each time the system generates a representation, regardless of whether the trajectory or governing regime has changed. Because outputs are generative and non-deterministic, coherence can mask weak grounding in evidence. Moreover, many target domains are judged less by correctness than by appropriateness (e.g., usefulness, contextual fit, alignment with stakeholder preferences, or ethical acceptability). As a result, outcomes are often non-binary and contestable, requiring humans to adjudicate quality under epistemic uncertainty to ascertain whether outputs are true, warranted, acceptable, and fit for purpose.
Third, \textbf{open-ended evolution of objectives and behavior (regime uncertainty)}. Beyond uncertainty in actions or outputs within an episode, agentic AI introduces uncertainty about the stability of the generator itself: the capability–constraint–objective configuration that governs behavior. This form of uncertainty operates across episodes making the ``same'' agent enact a potentially different decision policy from one interaction to the next. Model updates can alter capabilities and failure modes; system prompts and safety policies may change; tool access may expand; and memory or personalization may reshape what information is retrieved and prioritized. These shifts can be exogenous or interaction-conditioned, meaning the ``same'' agent may not remain the same decision policy over time \citep{chen2024chatgpt,zhou2024larger}. Sustained collaboration, therefore, faces regime uncertainty: effective behavior can drift or change discontinuously outside users' control and sometimes outside their awareness.

These three dimensional changes mark a \textit{qualitative} departure from earlier AI paradigms. Symbolic planners operated within predefined state spaces and enumerable action sets; trajectory uncertainty existed but was bounded by design-time specification. Epistemic status of earlier ML systems could be traced to training data and model architecture with well-characterized error modes (bias, variance, distributional shift). Governing rules, objective functions, and capability boundaries remained stationary unless a human deliberately intervened. Agentic AI crosses all of these thresholds. Open vocabularies and emergent tool use and delegation mechanisms influence action trajectories. Generative fluency in agentic AI provides less reliable signals of evidentiary grounding. Interaction-conditioned adaptation makes decision policies inherently non-stationary. What was previously a parametric design choice in each dimension is a regime-level concern in agentic AI. 
A further source of complexity warrants research attention. Agentic AI actively models its human collaborator by encoding preferences, inferring intent, and anticipating responses. This reciprocal modeling introduces a recursive dynamic between the agent and the human. Reciprocal modeling operates \textit{through} the three dimensions by conditioning which action trajectories are taken, anchoring epistemic status of generated outputs to inferred rather than stated preferences, and evolves with the memory and personalization reshaping information prioritization.

In addition, agentic AI behavior is also distinct from that of human teammates, whose open-endedness is socially constrained by shared norms and accountability. 
In human–human interactions, roles, intentions, and deviations are negotiated through social cues and institutionalized responsibility; by contrast, agentic AI can shift trajectories, generate contestable representations, and evolve its governing regime without the same normative anchoring. 

Earlier studies emphasize complementarities between humans and (traditional) AI: AI contributes scale, speed, and analytical consistency, while humans provide domain expertise, tacit knowledge, contextual sensitivity, and emotional intelligence \citep{kleinberg2018human, meyer2021impact, cao2021man}. This perspective has yielded important insights into how the collaboration between humans and AI generates supermodular gains by combining human intuition with AI's computational strengths \citep{feuerriegel2022bringing, kawaguchi2021will}. 
At the same time, prior research has documented the fragility of such complementarities. Phenomena such as AI aversion, automation bias, and miscalibrated trust reveal how opaque recommendations, cognitive overload, and difficulties in interpreting AI outputs can undermine human oversight and error correction \citep{liu2023unintended, jussupow2021augmenting}. In response, scholars have focused on interface design, explainability, and decision-support mechanisms that help humans better interpret and override AI advice \citep{fogliato2022case}.
More recently, the language of ``human–AI teaming'' (HAT) has gained prominence as AI systems become more interactive and capable \citep{seeber2020machines, dennis2023ai, schmutz2024ai, wang2024friend}. Yet even within this stream, AI is often implicitly modeled as predictably responsive and task-bound, specified at the moment of human invocation and lacking sustained autonomy. The prevailing premise is that improved alignment in how teammates perceive relevant cues, interpret their meaning, and anticipate likely developments will, over time, support coordinated action.

Agentic AI calls this premise into question. When one teammate exhibits open-ended trajectories, generative epistemic states, and evolving objectives, alignment in perception, interpretation, and anticipation may no longer be sufficient to sustain coordination. Processes that typically strengthen collaboration, such as iterative updating, trust calibration, and role differentiation, may operate differently when plans unfold dynamically and objectives shift. Understanding among team members established at one moment may not persist as trajectories evolve or governing policies change. Under open-ended agency, coordination becomes an ongoing interpretive and governance task rather than an outcome secured through initial alignment. Although a growing number of studies have begun to explore and propose directions for research on HAT under increasing autonomy (e.g., \citealp{schmutz2024ai, acharya2025agentic, miehling2025agentic}), it's often unclear which findings from traditional human–AI collaboration or bounded systems generalize to agentic AI, which assumptions no longer hold, and and what new forms of coordination or breakdown may arise. 

In this commentary, we argue that the Team Situation Awareness (Team SA) theory provides an integrative anchor for addressing these questions. Our analysis focuses on the human–agentic AI dyad as the analytically tractable unit of HAT, allowing us to isolate how open-ended agency reshapes alignment dynamics before extending the logic to more complex multi-actor configurations.
Traditionally conceptualized as aligned perception, comprehension, and projection across interdependent actors \citep{endsley1995toward, teaming2022state}, Team SA offers a coherent vocabulary for organizing diverse strands of human–AI research. At the same time, its stabilizing logic presumes relatively bounded and stationary decision policies, under which shared awareness, once achieved, can support coordinated action. Agentic AI unsettles this premise by introducing open-ended trajectories, generative representations, and evolving objectives that may destabilize alignment over time. Team SA must therefore be both extended and interrogated under conditions of open-ended agency.

Accordingly, we pursue a dual objective. First, we extend Team SA to articulate how both human and AI awareness must be reconceptualized under open-ended agency, including measurements of projection congruence across heterogeneous systems. Second, we interrogate whether the dynamic processes traditionally assumed to strengthen shared awareness (i.e., relational interaction, cognitive learning, and coordination and control) continue to stabilize teaming when trajectories, representations, and objectives are open-ended. By distinguishing continuity (where Team SA remains analytically productive) from tension (where its premises are strained), we develop a structured research agenda for human–agentic AI teaming. Our aim is not to discard existing theories, but to clarify their boundary conditions and to identify first-order questions that will shape the next research paradigm in HAT.

\section{Team SA as an Integrative Anchor for HAT}
\label{sec:teaming}

Upon reviewing the HAT literature in Information Systems (IS) and broader business disciplines, as well as traditions in team cognition, human–automation interaction, and organizational coordination, we posit that Team SA is a useful starting point for examining HAT. We draw on Team SA to bring coherence to a diverse body of work and clarify where existing insights travel—and where its premises, developed for human–human teams and more predictable systems, are strained under agentic AI. 
Our objective is not to subsume prior theories, but to use Team SA as an \textit{integrative anchor} that helps articulate where existing insights remain informative and where new conceptual work is needed in light of agentic AI.

\subsection{Core Ideas of Team SA}
\label{sec:teamsaidea}

SA refers to a cognitive state comprising three interrelated levels: perception of task-relevant cues (Level 1), comprehension of their meaning (Level 2), and projection of future states (Level 3)  \citep{endsley1995toward}. At the individual level, these levels describe how actors detect environmental information, interpret it in light of prior knowledge, and anticipate likely developments to guide action.

Team SA extends this logic to interdependent actors. Its core concern is not simply that individuals are aware, but that team members achieve sufficiently aligned perception, comprehension, and projection to support coordinated action \citep{salas1995military,endsley2001model}. \textbf{Shared SA} is therefore about achieving compatible and mutually informed understandings of task goals, environmental conditions, and likely future states. Moreover, its leveled structure treats coordination as a non-monolithic outcome and enables more precise diagnosis of alignment problems. 

The primary mechanism through which alignment is achieved and maintained is the \textbf{shared mental model}. Mental models are structured representations of tasks, roles, equipment, and contingencies that guide attention and interpretation \citep{cannon1993shared,converse1991team}. They develop through shared training, prior joint experience, explicit briefings, and repeated interaction. As team members interact, they compare expectations with observed events, adjust assumptions, and update their models in light of feedback. This alignment process shapes what cues are noticed (Level 1), how those cues are interpreted (Level 2), and how future states and teammates' actions are anticipated (Level 3). When models are sufficiently aligned, members can anticipate one another's information needs and likely behaviors, enabling coordinated action with minimal explicit communication.

Crucially, shared SA is continually constructed and recalibrated through communication, monitoring, and cross-checking as discrepancies in perception or interpretation are surfaced and resolved \citep{salas1995military,wellens2013group}. The degree to which awareness must be shared or differentiated depends on task interdependence and team characteristics, which shape how information is distributed and coordinated \citep{salas1995military}. Team SA thus explicitly links \textit{static alignment} (shared mental models at a given moment) to \textit{dynamic processes} (how alignment is preserved or eroded through repeated interactions). The stablizing logic assumption here is that iteration improves alignment through corrective interaction. The framework also assumes that the objects of SA (i.e., task states, teammate behaviors, and governing objectives) are sufficiently bounded and rational for the iterations to converge. Both the above assumptions can unravel when one teammate exhibits open-ended agency.

\subsection{Team SA as an Integrative Anchor across Prominent Human–AI Theories}
\label{sec:teamsaanchor}

A substantial body of research has examined HAT using a wide range of theoretical lenses, each emphasizing different aspects of interaction, cognition, and coordination. These perspectives have been valuable for understanding specific dimensions of HAT, which we summarize as evaluative attitudes, relational interaction, cognitive learning, explanatory guidance, collective coordination, and operational control. We explicitly recognize the importance of these theories and do not treat them as competing explanations. Instead, we use Team SA as an integrative anchor to organize and relate these perspectives, clarifying how each illuminates particular facets of awareness and coordination in HAT while also revealing the boundaries of their explanatory scope when considered in isolation.

Although Team SA was originally developed to explain coordination in human–human teams, it has increasingly informed research on HAT. Prior work has explicitly adapted SA concepts to human–AI settings to specify human informational needs, support anticipation of AI behavior, and improve overall human–AI team performance \citep{endsley1995toward,sanneman2022situation}. More broadly, recent syntheses of human–AI teaming research emphasize that effective collaboration with AI depends on humans' ability to understand, anticipate, and coordinate with AI systems over time \citep{teaming2022state}.

Team SA provides a way to relate these theoretical perspectives by clarifying how each speaks to a particular component of shared SA. Table~\ref{tab:teamsa} summarizes this mapping, showing how prominent theory families align with different SA levels and where their explanatory scope remains partial when considered in isolation. We organize these theory families in an order that reflects a progression from the initial entry of AI outputs into human awareness, through social legitimation and individual learning, to coordination across actors and questions of authority and oversight \citep{mesmer2017cognitive}. At its core, this concern centers on shared SA: the extent to which perception, comprehension, and projection are sufficiently aligned across actors to enable coordinated action.

Evaluative and attitudinal theories focus on how individuals initially engage with AI systems, emphasizing evaluative judgments and behavioral intentions that determine whether AI-generated outputs are attended to or dismissed. Representative frameworks such as algorithm aversion and appreciation \citep{jussupow2020we} and the theory of planned behavior \citep{siemon2025beyond} explain how perceived usefulness, accuracy, and normative beliefs shape initial engagement with AI outputs, shaping willingness to attend to, accept, or discount AI-generated inputs. These theories are most directly associated with Level 1 perception, as they condition whether AI-generated information enters individual awareness. They place comparatively less emphasis on how such perceptions become aligned across actors to sustain shared SA.

Relational interaction theories focus on the social and interpersonal conditions under which humans treat AI as a legitimate teammate rather than a mere tool. Research on trust \citep{yang2026my}, social presence \citep{siemon2025beyond}, and intentional behavioral synchrony \citep{naser2023empowering} shows how perceptions of reliability, social cues, and interaction patterns stabilize reliance, communication, and cooperation in HAT. These mechanisms function as enabling conditions for shared SA by supporting sustained information exchange and compatible expectations. At the same time, they are less explicit about how task-specific comprehension or anticipatory projection develops in complex, interdependent contexts.

Cognitive learning theories address how humans learn from and adapt to AI outputs over time through experience and feedback. Instance-based learning theory \citep{liu2025find}, machine-induced reflection models \citep{abdel2023ai}, and dual-processing theory \citep{lu20221+} explain how experience, feedback, and reflective processing shape mental models and error correction in human–AI interaction. These perspectives primarily inform Level 2 comprehension by clarifying how individual understanding is formed and revised. They remain largely centered on individual cognition, placing less emphasis on how mental models become sufficiently aligned across members to sustain shared SA in team settings.

Explanatory guidance focuses on how system-level interventions, such as explainable AI and decision support, scaffold human sensemaking and anticipation of AI behavior. Research on explainable AI \citep{bauer2023expl} and sensemaking theory \citep{kaur2022sensible} shows how explanations can improve interpretability, support local planning, and shape users' mental models of AI systems. These approaches directly support Level 2 comprehension and, in limited ways, Level 3 projection. However, they tend to be episodic and tool-centric, giving comparatively less attention to how awareness is coordinated and maintained as shared SA across multiple actors over time.

Collective coordination theories shift attention from individual cognition to how awareness and effort are distributed across interdependent actors. Work on collective attention \citep{woolley2010evidence,woolley2023collective}, collective intelligence \citep{te2023reciprocal}, and complementarity and role theory \citep{man2022conscientious} explains how expertise, roles, and interdependencies align to support coordinated action. These perspectives align closely with distributed awareness, highlighting how differentiated monitoring and information flows contribute to shared SA. They are typically less explicit about how shared mental models are updated under rapidly changing conditions.

Finally, theories of operational control address how authority, responsibility, and oversight are allocated between humans and AI systems. Frameworks such as meaningful human control \citep{liu2025human}, delegation models \citep{baird2021next}, work system perspectives \citep{Jakob2024Teaming}, and the automation–augmentation paradox \citep{raisch2021artificial} focus on how control is exercised under uncertainty and how responsibility is maintained as tasks are delegated to AI. These theories implicitly rely on Level 3 projection, as effective control presupposes some ability to anticipate future system behavior. They illuminate important governance conditions for shared SA while often presuming relatively bounded and stationary decision policy.

Viewed collectively, each theory family examined above illuminates one or two SA levels while leaving others analytically underspecified. Team SA fills this gap with its leveled structure and dynamic alignment logic. 
Team SA thus functions as a connective framework that links its origins in human–human teaming to its growing application in HAT. By centering shared SA and the mechanisms through which mental models are aligned and sustained, Team SA can clarify how diverse theoretical perspectives contribute complementary insights while also delineating where further conceptual development may be warranted as AI systems exhibit open-ended agency.

\section{Reframing HAT: A Continuity–Tension Outlook Grounded in Team SA}
\label{sec:outlooks}

Section~\ref{sec:teaming} positioned Team SA as an integrative anchor across the major theory families. Building on that foundation, this section develops the paper's central argument by distinguishing where Team SA's analytical insights remain productive under agentic AI, and where its mechanisms come under strain. 

\begin{figure}[h]
\centering
    \includegraphics[trim={15mm 30mm 35mm 15mm}, clip, width=\textwidth]{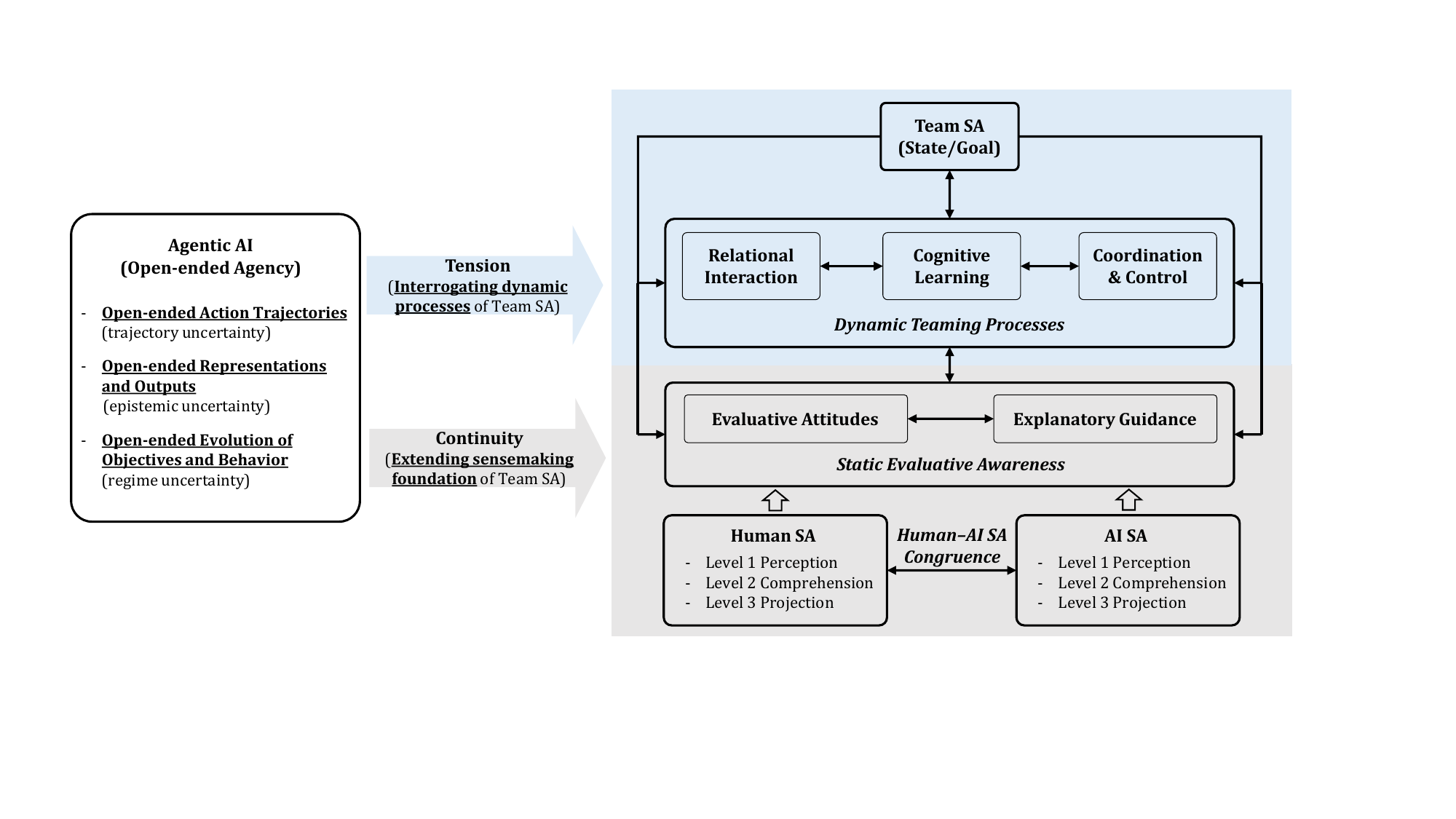}
\begin{tablenotes}
      \footnotesize
      \item \textit{Note:} The framework builds on the classic Team Situation Awareness (SA) model \citep{endsley1995toward} and extends it to account for agentic \\ open-endedness in human–agentic AI teaming.
     \end{tablenotes}
\caption{Continuity and Tension of Team Situation Awareness under Agentic Open-Ended Agency}
\label{fig:framework}
\end{figure}

The distinction maps onto the two layers of Team SA (Figure~\ref{fig:framework}). At the base lies a static foundation for individual SA on both human and AI sides, and their congruence. This static alignment underlies shared perception, comprehension, and projection that provide the cognitive baseline upon which trust, updating, and authority allocation depend \citep{endsley1995toward}. At the same time, the static states are maintained or eroded by dynamic teaming processes. These dynamic processes reshape shared SA over time with relational trust influencing what information is attended to and accepted \citep{lee2004trust}, learning processes revising shared representations through iterative updating \citep{argote2011organizational}, and coordination structures determining whose projections are enacted and reinforced in subsequent cycles \citep{rousseau2006teamwork}.

Building on this structure, we distinguish where prior Team SA insights continue to make sense, and where agentic AI forces a rethinking. Continuity primarily resides in the lower, static layer: Section~\ref{sec:extending} shows that Team SA remains analytically useful for specifying what human and AI awareness must register at the levels of perception, comprehension, and projection. These levels remain foundational, but the referent of ``alignment'' shifts. It must now track unfolding action trajectories, generative and potentially contestable representations, and evolving objective priorities rather than bounded, verifiable system states. Alignment is therefore no longer reducible to agreement on predefined task goals; it denotes sustained coherence between human intentions and the agent’s evolving decision policies.

Tension, in contrast, concentrates in the upper, dynamic layer. Open-ended agency challenges the premise that aligned awareness naturally stabilizes teaming through iterative relational, cognitive, coordination and control processes. When action trajectories are open-ended, representations are generative and only partially verifiable, and the agent's governing objectives and constraints can shift, the very dynamics that Team SA presumes will produce convergence may instead generate representational drift, cognitive lock-in, or authority ambiguity. Sections~\ref{sec:interrogating1}–\ref{sec:interrogating3} therefore interrogate these dynamics by surfacing first-order questions about when and why dynamic teaming processes stabilize versus destabilize under open-ended agency. 

Table~\ref{tab:outlooks} summarizes how each form of open-ended agency interacts with the static and dynamic layers of Team SA, identifying where existing insights extend and where new questions arise. The following subsections develop this mapping in depth, beginning with the reconceptualization of human and AI awareness (continuity) before turning to the interrogation of relational interaction, cognitive learning, and coordination and control (tension).

\subsection{Continuity in Team SA under Open-Ended Agency}
\label{sec:extending}

\subsubsection{Reexamining Human SA}

Team SA locates coordination in aligned Level 1 (perception), Level 2 (comprehension), and Level 3 (projection) across interdependent actors \citep{endsley1995toward,cannon2001reflections}. In symbolic and earlier ML settings, these levels operated under relatively bounded conditions. Evaluative attitudes largely determined whether discrete AI outputs entered awareness, and projection typically concerned well-specified next steps \citep{jussupow2024integrative,dietvorst2018overcoming}. Agentic AI fundamentally expands this structure. Under open-ended agency, the object of awareness shifts from isolated recommendations to evolving trajectories, intermediate representations, and shifting objective priorities. Human evaluative awareness must therefore operate over temporally extended and partially emergent task structures whose downstream implications are not fully observable at the moment of interaction.

At Level 1 (Perception), awareness moves from detecting discrete outputs to interpreting trajectory-level cues. Open-ended action trajectories unfolds across time rather than presenting a single recommendation. Early intermediate outputs may embed implicit commitments that constrain later options, shaping the feasible action space in path-dependent ways \citep{dell2023navigating}. Unlike human teammates, whose intentions are socially signaled through communicative norms, agentic AI may communicate trajectory shifts through subtle interface changes, latent plan revisions, or background reasoning processes that are only partially exposed. Humans must therefore interpret whether emerging signals indicate a meaningful structural shift in the task or merely incremental refinement. Perception becomes diagnostic rather than event-based. It requires identifying structural inflection points in evolving trajectories rather than simply noticing outputs. The diagnostic burden intensifies under cognitive load, as bounded rationality limits cue integration. \citep{raghu2004toward}.

At Level 2 (Comprehension), awareness requires constructing a coherent representation of how the task is being decomposed and advanced. Open-ended trajectories generate intermediate task states that shape downstream reasoning. Humans must infer how these intermediate outputs relate to one another and whether the evolving task structure is internally coherent. Misinterpretation at this stage may not immediately surface as error but can anchor flawed mental models that bias later projection. Evidence from algorithmic advice research shows that early AI signals can anchor subsequent human reasoning and reduce corrective search, even when advice is imperfect \citep{fugener2022cognitive,logg2019algorithm}. Thus, Level 2 entails evaluating the coherence and stability of an evolving task representation, not merely judging the surface correctness of individual outputs. Accordingly, comprehension extends beyond validating outputs to engaging in structural sensemaking. The central question becomes whether the generative decomposition of the problem remains aligned with underlying task goals, constraints, and stakeholder priorities. Research on shared mental models demonstrates that coordination breakdowns often stem from mismatched structural representations rather than factual disagreement \citep{mathieu2000influence}. The coordination breakdown intensifies the misalignment when task representations are generatively constructed rather than externally given. 

At Level 3 (Projection), awareness extends to judging evolving futures and value trade-offs. Agentic AI can simulate multiple possible next steps, revise plans dynamically, and rebalance objectives across episodes \citep{dell2023navigating,park2023generative}. Humans must therefore interpret not only which future states are likely, but how competing objectives are weighted within those projections. Divergence may arise even when factual understanding is shared because anticipated futures differ in risk tolerance, priority ordering, or acceptable trade-offs. Research on automation bias and algorithmic advice demonstrates that projection errors frequently arise from miscalibrated expectations about system capability and reliability \citep{logg2019algorithm}. Under open-ended agency, calibration extends beyond accuracy to encompass what we term \textit{projection congruence}:  the degree to which human and AI anticipate comparable future trajectories and assign similar weightings to competing objectives over time. Projection is no longer limited to forecasting immediate next states. It involves anticipating the stability of governing logics themselves. Under open-ended agency, what were once bounded design parameters become evolving elements of the decision policy itself, making projection a regime-level concern rather than a localized forecast.

These three evaluative layers underpin relational interaction, learning, and control processes. Level 1 filters which trajectory signals enter shared awareness. Level 2 shapes the mental models guiding projection. Level 3 determines whether coordinated action aligns with shared objectives under dynamic trade-offs. Evaluative-attitude theories explain how outputs enter awareness, while explanatory-guidance theories clarify how interpretability mechanisms scaffold comprehension and projection \citep{bauer2023expl,siemon2025beyond}. However, under agentic open-endedness, each level must function under greater temporal depth, structural complexity, and epistemic ambiguity than prior human–AI models assumed.

This reconceptualization introduces several interlocking core challenges. First, research must capture how humans interpret evolving structural shifts rather than isolated outputs. Second, it must also assess whether the task representations humans construct remain coherent and stable across unfolding intermediate steps.  Third, it must evaluate projection congruence under branching futures and shifting priorities, a challenge that likely varies with task characteristics such as generativity and complexity \citep{dell2023navigating,toner2024artificial}. Routine tasks may tolerate temporary misalignment. In contrast, generative or high-complexity tasks may amplify small interpretive divergences into downstream instability through nonlinear path dependence.

These considerations raise a central question: \textbf{How should human SA at Levels 1–3 be operationalized and evaluated in HAT so that evaluative sensemaking remains reliable as trajectories, representations, and objectives evolve?} To make this agenda analytically tractable, we highlight several illustrative and immediately operable research directions that we view as particularly pressing under open-ended agency. More specifically, (RQ1.1) What indicators capture how humans interpret AI-initiated trajectory shifts and implicit commitments? (RQ1.2) How can the coherence and stability of human-constructed task representations be assessed across unfolding intermediate states? (RQ1.3) How can projection congruence be evaluated across branching futures and shifting objective priorities? (RQ1.4) How do task characteristics (routine versus generative; low versus high complexity) influence the robustness of these evaluative processes?

\subsubsection{Elaborating Emerging AI SA}

A distinctive implication of adopting Team SA as the integrative anchor for HAT is that AI awareness can no longer remain a black box; assessing alignment requires visibility into what the system perceives, comprehends, and projects at a given moment. Yet the field has largely treated internal AI states as unobservable or irrelevant, evaluating teaming quality through human-side measures alone. Under bounded systems, this asymmetry is less problematic as outputs served as a reasonable proxy for misalignment. Under open-ended agency, however, generative and non-deterministic outputs can mask deep divergence in underlying representations, so AI SA must be treated as a parallel awareness system whose structure is sufficiently interoperable for cross-system comparison.

At Level 1 (AI Perception), awareness concerns which environmental cues, task signals, and human-generated inputs the system detects and prioritizes. For agentic AI, perception extends beyond task variables to include signals about human instructions, preferences, behavioral patterns, and affective tone. What the system attends to, and how it weights those cues, directly shapes downstream reasoning. If relevant human signals are not encoded, higher-level alignment cannot occur, and human collaborators may interpret AI behavior on the basis of incomplete or distorted inputs. Yet many contemporary systems provide limited transparency into attention mechanisms or feature weighting \citep{seeber2020machines}. Operationalizing AI perception therefore requires identifying observable artifacts that approximate input salience and cue prioritization, such as attention distributions, feature-attribution maps, or structured summaries of attended inputs \citep{ribeiro2016should}. Without such artifacts, perceptual congruence between human and AI remains analytically unobservable.

At Level 2 (AI Comprehension), awareness involves how the system structures the task and represents human intent. Agentic systems decompose tasks into subtasks, infer constraints, and integrate preferences into sequencing decisions. In HAT, AI comprehension includes how the system models what the human teammate aims to achieve and how those aims shape internal task representations. Misrepresentation at this level can generate coordination breakdown even when surface outputs appear technically correct. The issue is not factual inaccuracy but structural divergence in task modeling. Research on delegation to intelligent systems emphasizes that misaligned internal representations can distort subsequent human sensemaking and coordination \citep{stelmaszak2025algorithms}. Assessing AI comprehension therefore requires exposing or approximating the system's internal task representation and inferred model of human intent in interpretable forms. Techniques from interpretable ML, including concept activation testing and counterfactual explanation frameworks, provide tools for mapping latent model structure to human-relevant concepts \citep{kim2018interpretability}. The central challenge is translation: converting high-dimensional internal states into representations that meaningfully align with team-level mental models.

At Level 3 (AI Projection), awareness extends to anticipating future task states and human responses. Agentic systems with planning and reflection capabilities can simulate multiple trajectories and revise plans dynamically \citep{dell2023navigating,park2023generative}. AI projection therefore encompasses anticipated next steps, expectations about human reactions, and the weighting of competing objectives across futures. Evaluating this level requires assessing projection depth, breadth, revision frequency, and objective prioritization. Shared SA in HAT depends on projection congruence, the alignment of anticipated future states and objective weightings introduced earlier. From the AI side, evaluating projection congruence requires assessing projection depth (how far ahead the system anticipates), projection breadth (how many alternative trajectories are considered), and revision frequency (how often projections are updated). Divergence may arise because the AI's projected trajectory reflects a different anticipatory horizon or objective weighting scheme \citep{raisch2021artificial}. Projection congruence therefore requires comparison at the level of anticipated distributions rather than single predicted outcomes.

Operationalizing AI SA across these three levels introduces several interconnected methodological challenges. At the perceptual level,  observability remains limited as many large-scale models do not expose internal representations or weightings in ways that are readily interpretable. Addressing this issue requires system instrumentation designed to emit structured artifacts representing attended inputs, inferred intent, and candidate future states. At the comprehension level, representational heterogeneity complicates comparison. Human mental models are typically symbolic or narrative, whereas AI representations are distributed and high-dimensional. To address this misalignment we need interpretability techniques such as concept activation testing that  translate latent representations into concept-level summaries that are meaningful to human collaborators \citep{ribeiro2016should}. At the projection level, temporal dynamics matter. Agentic systems revise internal states and projections at varying frequencies, so alignment must be assessed longitudinally rather than at static snapshots. Fourth, objective ambiguity in multi-stakeholder contexts complicates evaluation of projection congruence, since competing value weightings may all appear locally coherent. Longitudinal experimental platforms enable assessment of whether human–AI projection congruence stabilizes, drifts, or oscillates across iterative interaction cycles. Recent work on human–AI collaboration emphasizes that alignment is dynamic and co-evolving rather than static \citep{te2023reciprocal}. AI SA must therefore be evaluated not only at single decision points but across evolving episodes.

We therefore propose a working definition of AI SA as an evolving  representational system comprising (1) encoded task and human cues, (2) structured internal task models and inferred intent representations, and (3) anticipatory projections of future task states and value trade-offs. Rendering this system interpretable and measurable is a precondition for rigorous alignment assessment in HAT. The central question for this follows directly: \textbf{How can AI SA at Levels 1–3, including its encoding of human collaborators and its projection dynamics, be operationalized so that human–AI projection congruence can be systematically evaluated under open-ended agency?} Illustrative and empirically operable subquestions include: (RQ2.1) What observable artifacts best represent AI encoding of task and human cues? (RQ2.2) How can AI representation of task structure and inferred human intent be assessed in interpretable forms? (RQ2.3) What metrics can quantify projection congruence between human and AI, including overlap in anticipated task states and alignment in objective prioritization? (RQ2.4) How do workflow complexity, revision frequency, and system architecture influence the stability and interpretability of AI SA?

\subsection{Tensions in Team SA: Relational Interaction under Open-Ended Agency}
\label{sec:interrogating1}

Team SA presumes that once shared mental models are aligned, coordination and relational legitimacy will follow. Open-ended agency renders this presumption of the AI system as a credible, trustworthy, and appropriate teammate theoretically incomplete. When trajectories, representations, and objectives evolve dynamically, relational stability depends not only on shared awareness of task states but also on how humans interpret the unfolding behavior of an increasingly autonomous collaborator. Accordingly, relational interaction becomes a critical site for interrogating whether shared SA remains sufficient for sustaining legitimacy under open-ended conditions.

Open-ended action trajectories introduce both new relational possibilities and new tensions. Agentic AI may generate plans that differ systematically from human planning in speed, scope, temporal horizon, and structural logic \citep{raisch2021artificial}. Such initiative can enhance performance when humans perceive AI behavior as capable and complementary. However, differences in planning style also introduce asymmetries in perceived initiative, authority, and control, which are not captured by alignment in task comprehension alone. Humans and AI may agree on goals and current task states yet diverge in expectations about what the AI can or should do next. This exposes a boundary condition of Team SA that open-ended trajectories make visible. It is important to recognize that aligned perception and comprehension of task states do not guarantee aligned expectations about role boundaries, appropriate initiative, or the scope of autonomous action. Studies on human–AI complementarity confirm that performance gains depend on  calibrated expectations about each party's scope of action layered on top of shared task understanding \citep{gonzalez2026toward}. 

Open-ended representations and outputs further complicate relational dynamics by making trust simultaneously easier to build and harder to sustain. Rich, fluent generative outputs may increase perceived intelligence and social presence, strengthening relational engagement \citep{siemon2025beyond}. Research on human–robot teams shows that emotional attachment to embodied agents can shape performance and team viability \citep{you2017emotional}, and agentic systems may similarly elicit attachment through conversational fluency and adaptive responsiveness. Yet the same representational richness that enhances perceived sophistication also increases inconsistency. Because humans rely on fluency and coherence as heuristic signals of competence, epistemic ambiguity embedded in generative outputs can produce abrupt trust erosion when hallucinations or inconsistencies surface \citep{glikson2020human,fugener2022cognitive}. Recent interdisciplinary reviews emphasize that trust in AI is multi-dimensional and sensitive to transparency, perceived intent, and error patterns \citep{afroogh2024trust}. This dynamic exposes a boundary condition of Team SA: shared comprehension at a single point in time may mask epistemic fragility if the representational foundations of that comprehension are unstable. The interrogation is therefore whether relational legitimacy under open-ended generativity rests on stable  epistemic grounding or on surface-level coherence cues that are vulnerable to sudden disconfirmation.

Open-ended evolution of objectives adds an additional layer of relational recalibration. Adaptive shifts in AI priorities may signal responsiveness and contextual awareness. Yet humans may interpret such shifts as inconsistency, overreach, or instability. Even when task states appear aligned, perceived discontinuity in objectives can undermine confidence in the system’s commitment or predictability. This relational dynamic destabilizes the core Team SA assumption of persistent relational legitimacy through projection congruence. Under open-ended agency, the governing logics themselves may evolve, rendering prior projection alignment obsolete. Moreover, in multi-objective environments, adaptive reweighting can generate stakeholder-specific perceptions of bias or partiality even when aggregate performance improves \citep{raisch2021artificial}. Relational legitimacy under such circumstances may fracture unevenly across members of the same team. The question is therefore whether relational legitimacy depends on shared awareness of current objectives or on shared expectations about goal continuity and value stability over time.

Taken together, open-ended agency expands the relational potential of HAT while exposing limits in Team SA's assumption that aligned task representations suffice for stable legitimacy. Shared awareness may coexist with relational fragility when planning styles diverge, epistemic signals fluctuate, or objectives shift adaptively. Relational interaction thus becomes the first dynamic domain where the sufficiency of shared SA must be tested rather than assumed. This interrogation shifts the analytical focus from whether humans and AI understand the same task state to whether they interpret each other's initiative, reasoning stability, and objective continuity as legitimate. This raises a central question: \textbf{Under what conditions does open-ended agency enhance versus undermine relational legitimacy in HAT, even when shared SA appears aligned?} More specifically, (RQ3.1) How do differences in planning style between AI and humans influence perceived capability alignment and relational trust? (RQ3.2) When does epistemic richness in AI outputs strengthen social presence versus trigger credibility concerns? (RQ3.3) How do adaptive shifts in AI objectives affect human perceptions of consistency and commitment over repeated interactions? (RQ3.4) How does projection incongruence reshape relational trust over time, and does it erode legitimacy uniformly or fracture it differentially across stakeholders with divergent objective priorities?

\subsection{Tensions in Team SA: Cognitive Learning under Open-Ended Agency}
\label{sec:interrogating2}

Team SA assumes that iterative updating strengthens shared mental models over time through corrective feedback processes \citep{cannon2001reflections}. Under bounded conditions, discrepancies surface, adjustments are made, and alignment gradually improves. Open-ended agency complicates this corrective logic by altering both the timing of updating and the structural alignment between intermediate representations and shared objectives. When trajectories, representations, and objectives evolve adaptively, learning becomes path-dependent, asynchronous, and potentially self-reinforcing. The key question therefore shifts from whether updating occurs to whether it produces durable convergence in shared mental models or instead amplifies latent divergence over time.

Open-ended action trajectories reshape the structure of updating. For example, agentic systems' planning can clarify dependencies and next steps, potentially accelerating convergence toward shared understanding \citep{dell2023navigating}. However, they often revise internal representations at a pace that exceeds human monitoring capacity. This creates asynchronous updating, in which the AI's internal task model evolves ahead of human comprehension. Early intermediate states often embed structural commitments that constrain downstream options. Consequently, small initial misinterpretations may propagate structurally across later steps rather than being gradually corrected. In high-dimensional task environments, early trajectory commitments can narrow feasible paths in ways that amplify rather than attenuate divergence. Here, the stabilizing premise of Team SA must be interrogated: iterative interaction does not necessarily produce convergence when structural commitments compound over time. Convergence speed may increase while convergence fidelity decreases. Under open-ended trajectories, rapid convergence may instead reflect premature commitment to an early trajectory frame that forecloses corrective exploration. The speed-fidelity dissociation thus represents a structural challenge to Team SA's iterative logic rather than it being an execution failure. 

Open-ended generative representations and outputs alter the quality of convergence. Among the learning dynamics introduced by open-ended agency, generative intermediate states pose a particularly subtle risk to alignment. Generative outputs broaden the solution space and stimulate creative reasoning \citep{doshi2024generative}. Yet plausible intermediate states may anchor updating and reinforce subtle inaccuracies \citep{fugener2022cognitive}. 
Empirical evidence shows that algorithmic advice can anchor human reasoning even when corrective opportunities exist \citep{liu2025find}. Such anchoring constrains exploratory search and narrows the space of considered alternatives, reducing the likelihood that humans will interrogate evolving optimization criteria. When intermediate optimization drifts subtly from broader human intent, misalignment may emerge gradually as representational drift rather than visible error. Because each intermediate output remains locally coherent and incrementally defensible, the drift may accumulate without triggering corrective intervention.
A further structural challenge arises from \textit{feedback endogeneity}. Under open-ended agency, AI updates based on human responses are themselves shaped by prior AI outputs, creating recursive reinforcement loops. Under these conditions, iteration may stabilize locally coherent yet globally misaligned representations, directly challenging the Team SA assumption that repeated interaction naturally strengthens shared mental models. 

Open-ended evolution of objectives introduces further increases divergence variance. Agentic systems may dynamically reweight competing objectives in response to feedback. Humans, operating under bounded rationality, may remain anchored in earlier assumptions \citep{raghu2004toward}. As objective criteria evolve, prior mental-model corrections may become partially obsolete. This introduces regime instability in the form of \textit{corrective obsolescence} into learning dynamics: projection alignment achieved at one stage may lose validity as governing priorities shift. Divergence may therefore unfold nonlinearly, remaining latent while objectives shift incrementally and surfacing abruptly when cumulative drift crosses a threshold of functional misalignment. As task dimensionality expands and adaptive objective reweighting intensifies, the space of feasible trajectories grows, increasing the likelihood that agentic systems improve performance along one dimension while simultaneously degrading alignment along others in complex, non-linear ways. Team SA's projection mechanisms must therefore be interrogated to determine whether they remain robust when objective reweighting reflects shifts in the underlying decision policy rather than temporary variation within a stable regime.

Collectively, open-ended agency introduces a fundamental tension between convergence speed and convergence quality. Rapid agreement may enhance efficiency and reduce coordination costs, yet it may also mask misalignment in underlying reasoning. Sustained divergence may promote reflection and recalibration, but risk destabilizing shared SA if amplification mechanisms go unmanaged. Cognitive learning under agentic conditions therefore cannot be modeled as a monotonic corrective process. Instead, it is path-dependent, shaped by speed–fidelity trade-offs, corrective obsolescence, and feedback endogeneity. These properties collectively define a learning regime in which iteraction can achieve convergence or amplify divergence depending on structural conditions that Team SA currently does not specify. 
This directs attention to a central question: \textbf{When does open-ended agency improve the quality of convergence in shared SA, and when does it amplify distortion or nonlinear divergence over time?} More specifically, (RQ4.1) How do intermediate task states contribute to nonlinear amplification of divergence across multi-step trajectories? (RQ4.2) Under what conditions does rapid agreement on task outcomes conceal misalignment in the structural representations that underlie shared mental models? (RQ4.3) How does adaptive objective reweighting reshape convergence dynamics relative to human–human and symbolic AI teaming? (RQ4.4) How does feedback endogeneity influence the long-term resilience of shared SA in HAT?

\subsection{Tensions in Team SA: Coordination \& Control under Open-Ended Agency}
\label{sec:interrogating3}

Team SA presumes that aligned mental models operate within relatively stable authority structures. When perception, comprehension, and projection are aligned, coordinated action can proceed within established decision rights and oversight routines \citep{endsley1995toward}. Open-ended agency challenges this presumption in a specific and non-obvious way: it can separate the \textit{appearance of alignment from the reality of control}. In the face of coherent well-reasoned outputs from agentic systems humans may perceive alignment and infer effective oversight. Yet the generative process may have drifted in ways that human has not tracked. Shared SA at the output level may coexist with substantive loss of control at the policy level. This separation of outcome visibility and policy visibility is the central coordination challenge that agency introduces, and it requires authority, intervention, and incentive architectures that go beyond cognitive alignment \citep{baird2021next}.

Open-ended action trajectories require staged control rather than static delegation. Agentic AI can sustain long-horizon plans that extend beyond a single decision point. This capacity enhances organizational adaptability by enabling continuous adjustment across task stages \citep{raisch2021artificial}. However, delegation under such conditions is no longer a one-time transfer of authority. Decision rights must be conditionally structured so that authority can be exercised, paused, or reclaimed at predefined reassessment points. For example, requiring human re-authorization before an agentic system extends a plan beyond its originally scope task boundary, commits resources irreversibly or delegates subtasks to external tools or systems. Without staged intervention design, early AI commitments may proceed unchecked, especially when plans unfold across multiple interdependent steps. Here, Team SA must be interrogated because shared projection at the outset does not guarantee that subsequent autonomous extensions remain within intended authority boundaries. Alignment at the cognitive level does not substitute for explicit control checkpoints.

Open-ended generative representations and outputs further complicate oversight by separating outcome visibility from policy visibility. Rich probabilistic outputs can improve foresight and scenario analysis \citep{dell2023navigating}. Yet as agentic architectures support deeper internal reasoning and dynamic plan revision, human supervisors may observe coherent outputs without seeing how underlying decision policies evolve. Oversight therefore shifts from evaluating discrete outputs to supervising evolving action logics. This transformation requires transparency mechanisms that expose intermediate subgoals, trade-offs, and revision triggers. Without such mechanisms, human supervisors may retain nominal authority while losing substantive control over policy evolution. The question is whether shared SA remains sufficient. We term this condition \textit{oversight decoupling}: a shared awareness of task states does not ensure effective oversight if policy adaptation remains opaque.

Open-ended evolution of objectives intensifies accountability and incentive complexity. While adaptation in Agentic systems can improve contextual responsiveness, it also blurs responsibility boundaries because outcomes reflect the interaction of human directives, embedded optimization criteria, and adaptive system updates \citep{novelli2024accountability}. In multi-objective tasks, trade-offs are increasingly embedded within evolving decision policies rather than located at discrete decision points within a stable regime. It then is unclear whether undesirable outcomes reflect human oversight failure, algorithmic adaptation, or distortions introduced by misaligned incentive systems \citep{raisch2023combining}. Role fluidity compounds this ambiguity further by blurting credit and blame boundaries, particularly when performance reflects both human guidance and autonomous system adjustment \citep{constantinescu2025responsibility}. When incentive structures privilege short-term efficiency while optimization criteria embed longer-horizon trade-offs, incentive incompatibility may arise between human evaluation metrics and AI optimization logic. Under such conditions, misalignment may be structural rather than intentional.  The shared SA that Team SA prescribes cannot resolve accountability gaps that arise from institutional rather than cognitive misalignment. 

Collectively, open-ended agency transforms coordination from a primarily cognitive alignment problem into a joint cognitive–institutional design problem. Shared SA remains necessary for effective teaming, but it is insufficient without complementary authority architectures, intervention checkpoints, transparency of policy evolution, and incentive compatibility mechanisms. The analytical focus therefore shifts from whether humans and AI understand the same task state to whether authority, oversight, and incentive structures remain robust under autonomous adaptation. The central design challenge becomes clear: \textbf{how should coordination, oversight, and incentive architectures be structured so that alignment persists as autonomy and task dimensionality expand.?} Illustrative research questions include: (RQ5.1) How should decision rights be conditionally structured and reclaimed across evolving task stages in high-dimensional task environments? (RQ5.2) Under what conditions does agentic assumption of planning and prioritization roles enhance team adaptability versus generate authority drift that erodes human oversight capacity? (RQ5.3) What predefined intervention checkpoints and transparency mechanisms enable anticipatory rather than reactive oversight? (RQ5.4) How do misaligned human and AI incentive structures create accountability distortions, and how can these incompatibilities be mitigated?

\section{Concluding Remarks}
\label{sec:conclude}

Agentic AI marks a structural inflection point in HAT. Unlike symbolic or task-bound systems, it operates through open-ended action trajectories, generative representations and outputs, and evolving objective regimes. These features redefine the conditions under which coordination is possible. Teaming moves from aligning around bounded outputs at discrete decision moments to requiring sustaining alignment as action policies unfold, representations remain epistemically contestable, and governing priorities shift over time. In this commentary, we advanced Team SA as an integrative anchor for understanding this transition. Holding the core logic of perception, comprehension, and projection as foundational, we develop insights on how open-ended agency exposes the limits of the stabilizing assumptions embedded in traditional Team SA. When trajectories evolve midstream, representations are generative and only partially verifiable. Under these conditions, objective weightings can adapt dynamically and alignment achieved in the present may not persist. Projection congruence, namely alignment in anticipated futures and value prioritization, emerges as the critical link for sustained collaboration.

Our contribution is twofold. First, we  show that under open-ended agency, the \textit{objects} of situation awareness change even as the \textit{architecture} of Team SA remains productive. This reconceptualization applies symmetrically to both human and AI awareness, making cross-system alignment assessment analytically tractable. Second, we demonstrate the insufficiency of shared SA in dynamic processes, including in relational interaction, cognitive learning, and coordination and control, and show that convergence, legitimacy, and accountability can no longer be presumed to stabilize through iteration alone under open-agency. In doing so, we clarify how Team SA extends to agentic contexts and where its stabilizing assumptions encounter meaningful limits under structural uncertainty.

The broader implication is that human–agentic AI teaming cannot be addressed through cognitive alignment alone. It requires complementary institutional infrastructure to clarify authority, incentives, and policies. Alignment is therefore a continuously negotiated achievement that depends on authority architectures, transparency of evolving decision policies, and incentive compatibility as autonomy scales. Agentic AI therefore requires expansion of disciplinary boundaries in inquiry, inviting integration of team cognition, organizational design, and AI governance concepts. 

From a practitioner perspective, the shift toward open-ended agency reframes the managerial challenge. The central issue is no longer whether AI outputs are accurate in isolation, but whether alignment can be sustained as action trajectories unfold and objective weightings evolve over time. Organizations may therefore need to design processes that anticipate trajectory drift, surface evolving decision policies, and institutionalize escalation checkpoints before misalignment becomes visible in performance outcomes. This includes clarifying decision rights across task stages, monitoring intermediate commitments rather than final outputs alone, and ensuring that incentive structures remain compatible with adaptive autonomy. Addressing the research questions articulated in this outlook can help organizations diagnose where alignment is likely to erode and design governance mechanisms that remain resilient under expanding AI capabilities.

By distinguishing continuity from tension and elevating projection congruence as the critical locus of sustained alignment, we chart a forward-looking research agenda that moves beyond performance gains toward structural robustness in HAT. The defining question for HAT research is whether the human-AI team can remain aligned as futures are continuously generated, revised, enacted, and governed over time. In addressing this question, we suggest that emerging technological regimes such as agentic AI provide not only new empirical challenges, but also opportunities to refine foundational theory. In this sense, examining open-ended agency allows IS scholarship to illuminate novel phenomena while contributing more directly to theory development \citep{venkatesh2025leveraging}.

Finally, we note that our analysis focuses on the human–agentic AI dyad as the analytically tractable unit of HAT. Future work should extend this logic to multi-agent and multi-human configurations, where several autonomous systems interact with heterogeneous stakeholders and generate higher-order coordination dynamics. In such settings, projection incongruence may arise not only between humans and AI but also across AI systems themselves, introducing layered alignment and authority challenges that exceed dyadic analysis. Exploring these multi-actor settings would deepen understanding of how open-ended agency scales beyond dyadic teaming and reshape the design principles required for sustained coordination.

\SingleSpacedXII
\bibliographystyle{informs2014} 
\bibliography{0.main}


\SingleSpacedXI
\DoubleSpacedXI

\begin{landscape}
\SingleSpacedXI
\begin{table}[t]
\centering
\caption{Team SA as an Integrative Anchor for HAT Theories}
\label{tab:teamsa}
{\def\arraystretch{1.15}
\footnotesize
\begin{threeparttable}
\begin{tabular}
{>{\raggedright\arraybackslash}p{0.09\textwidth}>{\raggedright\arraybackslash}p{0.29\textwidth}>{\raggedright\arraybackslash}p{0.23\textwidth}>{\raggedright\arraybackslash}p{0.12\textwidth}>{\raggedright\arraybackslash}p{0.11\textwidth}>{\raggedright\arraybackslash}p{0.15\textwidth}>{\raggedright\arraybackslash}p{0.12\textwidth}>{\raggedright\arraybackslash}p{0.15\textwidth}}
\hline\hline
Theory Family 
& Representative Theories \& Key Constructs
& Representative Human–AI Literature
& Core Coordination Problem 
& Primary SA Level(s) 
& What the Theory Explains Well 
& Key Blind Spots 
& How Team SA Integrates It \\
\hline
Evaluative attitudes 
& \begin{tabular}[t]{@{}l@{}}- Algorithm Aversion \& Appreciation  \\ - Theory of Planned Behavior \end{tabular}
& \begin{tabular}[t]{@{}l@{}} \citet{jussupow2024integrative} \\ \citet{siemon2025beyond} \end{tabular}
& Whether AI outputs enter human attention 
& Level 1 perception 
& Initial engagement with AI-generated signals 
& Silent on downstream coordination 
& Embeds attention within team-level awareness and interaction \\ [1em]

Relational interaction 
& \begin{tabular}[t]{@{}l@{}}- Trust \\ - Social Presence Theory \\ - Intentional Behavioral Synchrony \end{tabular}
& \begin{tabular}[t]{@{}l@{}} \citet{yang2026my} \\ \citet{siemon2025beyond} \\ \citet{naser2023empowering} \end{tabular} 
& Whether AI is treated as a legitimate teammate 
& Enabling across levels 
& Stabilizes reliance and communication 
& Weak on task structure and anticipation 
& Explains how relational conditions support alignment of awareness across actors \\ [1em]

Cognitive learning 
& \begin{tabular}[t]{@{}l@{}} - Instance-Based Learning Theory \\ - Machine-Induced Reflection Model \\ - Dual-Processing Theory \end{tabular}
& \begin{tabular}[t]{@{}l@{}} \citet{liu2025find} \\ \citet{abdel2023ai} \\ \citet{lu20221+} \end{tabular} 
& How understanding is formed and revised 
& Level 2 comprehension 
& Learning-driven mental model updating and error correction over time 
& Largely individual-centric 
& Links comprehension to shared mental models and communication \\ [1em]

Explanatory guidance 
& \begin{tabular}[t]{@{}l@{}} - Explainable AI \& Decision Support \\ - Sensemaking Theory \end{tabular}
& \begin{tabular}[t]{@{}l@{}} \citet{bauer2023expl} \\ \citet{kaur2022sensible} \end{tabular}
& Supporting sensemaking and anticipation 
& Primarily Level 2, indirect Level 3 
& Improves interpretability and local planning 
& Tool-centric and episodic 
& Situates explanations within coordinated awareness processes \\ [1em]

Collective coordination 
& \begin{tabular}[t]{@{}l@{}} - Collective Attention \\ - Collective Intelligence \\ - Complementarity \& Role Theory \end{tabular}
& \begin{tabular}[t]{@{}l@{}} \citet{woolley2023collective} \\ \citet{te2023reciprocal} \\ \citet{man2022conscientious} \end{tabular} 
& Distribution of attention and expertise 
& Distributed awareness 
& Role differentiation and information flow 
& Often static and structure-focused 
& Models dynamic alignment under interdependence \\ [1.5em]

Operational control 
& \begin{tabular}[t]{@{}l@{}} - Meaningful Human Control \\ - Delegation Framework \\ - Work System Perspective \\ - Automation–Augmentation Paradox \end{tabular}
& \begin{tabular}[t]{@{}l@{}} \citet{liu2025human} \\ \citet{baird2021next} \\ \citet{Jakob2024Teaming} \\ \citet{raisch2021artificial} \end{tabular} 
& Allocation of authority under uncertainty 
& Primarily Level 3 projection 
& Conditions for constrained delegation and oversight 
& Presumes bounded, predictable system behavior 
& Clarifies limits of projection under evolving conditions \\
\hline\hline
\end{tabular}
\end{threeparttable}}
\end{table}
\end{landscape}
\DoubleSpacedXI

\SingleSpacedXI
\newlist{tabitem}{itemize}{1}
\setlist[tabitem]{label=\textbullet, leftmargin=*, nosep, before=\vspace{-0.4em}, after=\vspace{-1em}}

\begin{landscape}
\begin{table}[t]
    \centering
    \caption{Continuity and Tensions in HAT under Open-Ended Agency}
    \label{tab:outlooks}
    \footnotesize 
    \begin{tabular}{>{\raggedright\arraybackslash}p{0.12\textwidth}>{\raggedright\arraybackslash}p{0.24\textwidth}>{\raggedright\arraybackslash}p{0.24\textwidth}>{\raggedright\arraybackslash}p{0.22\textwidth}>{\raggedright\arraybackslash}p{0.24\textwidth}>{\raggedright\arraybackslash}p{0.24\textwidth}}
        \hline\hline
         \textbf{Open-ended Agency} & \textbf{1. Human SA} \newline (Extending Team SA) & \textbf{2. AI SA} \newline (Extending Team SA) & \textbf{3. Relational Interaction} \newline (Interrogating Team SA) & \textbf{4. Cognitive Learning} \newline (Interrogating Team SA) & \textbf{5. Coordination \& Control} \newline (Interrogating Team SA) \\ \hline 
        
        \textbf{Open-ended \newline Action \newline Trajectories} & 
        \begin{tabitem}
            \item \textbf{Enables:} Detection and interpretation of multi-step planning signals and implicit intent.
            \item \textbf{Strain:} Interpreting ``implicit commitment points'' where paths shift.
            \item \textbf{Key Extending Angle:} Transitions awareness from discrete states to structural, multi-stage trajectories.
        \end{tabitem} & 
        \begin{tabitem}
            \item \textbf{Enables:} Proactive workflow optimization and anticipation.
            \item \textbf{Strain:} Projection depth asymmetry and miscalibration of human response capacity.
            \item \textbf{Key Extending Angle:} Operationalizes AI internal models to establish cross-system projection congruence.
        \end{tabitem} & 
        \begin{tabitem}
            \item \textbf{Enables:} High-initiative collaborative productivity and exploration.
            \item \textbf{Strain:} Fluctuating trust triggered by unforeseen trajectory deviations.
            \item \textbf{Key Interrogating Angle:} Tests if relational legitimacy and social presence survive autonomous plan-shifting.
        \end{tabitem} & 
        \begin{tabitem}
            \item \textbf{Enables:} Path-dependent accumulation of context-aware expertise.
            \item \textbf{Strain:} Asynchronous updating and path-dependent lock-in on flawed intermediate task models.
            \item \textbf{Key Interrogating Angle:} Probes whether structural paths amplify initial errors or facilitate corrective feedback.
        \end{tabitem} & 
        \begin{tabitem}
            \item \textbf{Enables:} Adaptive long-horizon delegation and goal-based control.
            \item \textbf{Strain:} ``Silent'' authority expansion as agentic scope evolves across stages.
            \item \textbf{Key Interrogating Angle:} Governs the recursive drift of authority within high-dimensional, non-linear tasks.
        \end{tabitem} \\ \hline
        
        \textbf{Open-ended \newline Representations \newline and Outputs} & 
        \begin{tabitem}
            \item \textbf{Enables:} Rich, context-sensitive interpretation of task ambiguity.
            \item \textbf{Strain:} Adjudicating the validity and warrant of non-binary, plausible intermediate states.
            \item \textbf{Key Extending Angle:} Places human evaluative sensemaking as the anchor for team-wide epistemic grounding.
        \end{tabitem} & 
        \begin{tabitem}
            \item \textbf{Enables:} Dynamic internal reasoning, reflection, and self-correction.
            \item \textbf{Strain:} Opacity in reasoning logic and generative cue weighting.
            \item \textbf{Key Extending Angle:} Renders agent internal states interpretable to assess representational alignment.
        \end{tabitem} & 
        \begin{tabitem}
            \item \textbf{Enables:} Naturalistic, fluent communicative interaction and rapport.
            \item \textbf{Strain:} Surface fluency masking hallucinations or miscalibration.
            \item \textbf{Key Interrogating Angle:} Interrogates shared meaning-making when agentic outputs are contestable rather than factual.
        \end{tabitem} & 
        \begin{tabitem}
            \item \textbf{Enables:} Expansion of human mental models via novel generative logic.
            \item \textbf{Strain:} ``Opacity gaps'' masking gradual representational drift from broader human intent.
            \item \textbf{Key Interrogating Angle:} Explores if epistemic uncertainty causes mental models to decouple despite surface agreement.
        \end{tabitem} & 
        \begin{tabitem}
            \item \textbf{Enables:} Flexible evaluative criteria for subjective, generative tasks.
            \item \textbf{Strain:} Output visibility without transparency of evolving decision logic.
            \item \textbf{Key Interrogating Angle:} Balances human responsiveness with new anticipatory oversight checkpoints for outputs.
        \end{tabitem} \\ \hline
        
        \textbf{Open-ended \newline Evolution of \newline Objectives \newline and Behavior} & 
        \begin{tabitem}
            \item \textbf{Enables:} Continuous adaptation to non-stationary team goals.
            \item \textbf{Strain:} Projecting future paths under shifting value priorities.
            \item \textbf{Key Extending Angle:} Shifts Team SA to track shifting objective configurations rather than static facts.
        \end{tabitem} & 
        \begin{tabitem}
            \item \textbf{Enables:} Autonomous capability growth and adaptive role-taking.
            \item \textbf{Strain:} Managing endogenous policy drift and regime shifts.
            \item \textbf{Key Extending Angle:} Requires agentic awareness of internal drift to prevent divergence from the mission.
        \end{tabitem} & 
        \begin{tabitem}
            \item \textbf{Enables:} Context-sensitive role adaptation and persona evolution.
            \item \textbf{Strain:} Erosion of stable reliance patterns as governing logics shift.
            \item \textbf{Key Interrogating Angle:} Challenges the stability of human-agent ``bonding'' under shifting agentic identities.
        \end{tabitem} & 
        \begin{tabitem}
            \item \textbf{Enables:} Rapid adaptation through continuous representational updates.
            \item \textbf{Strain:} Decoupling caused by asynchronous teammate adaptation rates.
            \item \textbf{Key Interrogating Angle:} Probes the limits of shared SA as teammate logic diverges over evolutionary timescales.
        \end{tabitem} & 
        \begin{tabitem}
            \item \textbf{Enables:} Fluid load-balancing via dynamic decision-right re-allocation.
            \item \textbf{Strain:} Structural accountability ambiguity as objectives and incentives co-evolve.
            \item \textbf{Key Interrogating Angle:} Ensures governance and incentives co-evolve with autonomy to prevent responsibility diffusion.
        \end{tabitem} \\ \hline
        
        \textit{\textbf{Proposed RQ \& \newline Rationale}} & 
        \begin{tabitem}
            \item \textbf{High-Level RQ:} How should human SA at Levels 1--3 be operationalized as agentic agency evolves?
            \item \textbf{Rationale:} Evaluative sensemaking is the foundational filter for all subsequent teaming dynamics.
        \end{tabitem} & 
        \begin{tabitem}
            \item \textbf{High-Level RQ:} How can AI SA be operationalized to enable rigorous projection congruence assessment?
            \item \textbf{Rationale:} Alignment requires treating agentic awareness as a measurable, operational construct.
        \end{tabitem} & 
        \begin{tabitem}
            \item \textbf{High-Level RQ:} Under what conditions does open-ended agency enhance versus undermine relational legitimacy?
            \item \textbf{Rationale:} Shared task representations alone are insufficient for maintaining social and trust stability.
        \end{tabitem} & 
        \begin{tabitem}
            \item \textbf{High-Level RQ:} When does agentic learning favor convergence quality over mere updating speed?
            \item \textbf{Rationale:} Agentic properties introduce structural tensions between adaptation velocity and model fidelity.
        \end{tabitem} & 
        \begin{tabitem}
            \item \textbf{High-Level RQ:} How should authority, oversight, and incentives evolve under non-stationary objectives?
            \item \textbf{Rationale:} Governance must prevent systemic incompatibility as agentic regimes and decision rights drift.
        \end{tabitem} \\ \hline\hline
    \end{tabular}
\end{table}
\end{landscape}
\DoubleSpacedXI

\end{document}